\documentclass[sigconf]{acmart}
\usepackage{subfigure}

\AtBeginDocument{%
  }

\copyrightyear{2022} 
\acmYear{2022} 
\setcopyright{acmcopyright}
\acmConference[CIKM '22]{Proceedings of the 31st ACM International Conference on Information and Knowledge Management}{October 17--21, 2022}{Atlanta, GA, USA}
\acmBooktitle{Proceedings of the 31st ACM International Conference on Information and Knowledge Management (CIKM '22), October 17--21, 2022, Atlanta, GA, USA}
\acmPrice{15.00}
\acmDOI{10.1145/3511808.3557623}
\acmISBN{978-1-4503-9236-5/22/10}

\begin{document}

\title{KSG: Knowledge and Skill Graph}

\author{Feng Zhao}
\affiliation{%
\institution{Westlake University}
\city{Hangzhou}
\state{Zhejiang}
\country{China}
}
\email{zhaofeng@westlake.edu.cn}

\author{Ziqi Zhang}
\affiliation{%
\institution{Westlake University}
\city{Hangzhou}
\state{Zhejiang}
\country{China}
}
\email{zhangzq20@mails.tsinghua.edu.cn}

\author{Donglin Wang}
\authornote{Corresponding author. \\
Acknowledgments: This work was supported by the National Science and Technology Innovation 2030 - Major Project (Grant No. 2022ZD0208800), and NSFC General Program (Grant No. 62176215).
}
\affiliation{%
\institution{Westlake University}
\institution{Westlake Institute for Advanced Study}
\city{Hangzhou}
\state{Zhejiang}
\country{China}}
\email{wangdonglin@westlake.edu.cn}



\begin{abstract}
  The knowledge graph (KG) is an essential form of knowledge representation that has grown in prominence in recent years. Because it concentrates on nominal entities and their relationships, traditional knowledge graphs are static and encyclopedic in nature. On this basis, event knowledge graph (Event KG) models the temporal and spatial dynamics by text processing to facilitate downstream applications, such as question-answering, recommendation and intelligent search. 
  Existing KG research, on the other hand, mostly focuses on text processing and static facts, ignoring the vast quantity of dynamic behavioral information included in photos, movies, and pre-trained neural networks. In addition, no effort has been done to include behavioral intelligence information into the knowledge graph for deep reinforcement learning (DRL) and robot learning. In this paper, we propose a novel dynamic knowledge and skill graph (KSG), and then we develop a basic and specific KSG based on CN-DBpedia. The nodes are divided into entity and attribute nodes, with entity nodes containing the agent, environment, and skill (DRL policy or policy representation), and attribute nodes containing the entity description, pre-train network, and offline dataset. KSG can search for different agents' skills in various environments and provide transferable information for acquiring new skills. This is the first study that we are aware of that looks into dynamic KSG for skill retrieval and learning. Extensive experimental results on new skill learning show that KSG boosts new skill learning efficiency.
\end{abstract}



\maketitle

\section{Introduction}
Knowledge Graph (KG), first announced by Google in 2012, is a popular and efficient model for knowledge representation that has attracted the interest of researchers in related fields \cite{2009dbpedia,2007yago,2014wikidata}. A knowledge graph is a knowledge base of information about entities (e.g., people and organizations) that represent the many entities and their relationships in a domain using a collection of subject-predicate-object triplets. Each triplet is also referred to as a fact. Nodes represent entities, while edges reflect relationships between things in a knowledge graph \cite{li2021learning,abu2021relational,li2021efficient,mezni2021context,2021knowledge,li2021graph,xue2021relation}.
Nowadays, along with the continuous development of intelligent information service applications, the knowledge graph has been widely applied to many fields, such as question-answering, recommendation, text generation and so on \cite{JensLehmann2015DBpediaA,li2020real,mezni2021temporal,zhao2021multimodal,zhao2021multi,zhao2022graph}.

However, there is a wealth of event information available in the world, such as the most recent news stories, which conveys dynamic and procedural knowledge. As a result, event-centric knowledge representation forms such as Event KG (EKG) have been a popular research topic \cite{2019SearchingNA,2019ATS,2020EventQAAD,JieWu2020,2021Event}. To address the needs of rapid retrieval and concise representation of event-related information, EKG can filter and structure information about events reported in texts \cite{2015CAE,2017EEGKB,ZhangKD20,zhang2022domain,zhang2022tree}. Meanwhile, in order to capture dynamic information of events, EKG considers action, participant, time, and location to extract event-event relations including temporal and causal relations \cite{2016BuildingEK,2018EventKGAM,2019ELGAE}. However, existing methods on KG or EKG usually focus on text processing and disregard dynamic behavior information, making it impossible to search for and understand the behavior or skills of humans and agents \cite{liu2022learn,tian2021independent,tian2021unsupervised,liu2022dara,liu2021unsupervised}.


In this paper, we present a novel concept of Knowledge and Skill Graph (KSG) to address above problems. Based on CN-DBpedia \cite{xu2017cn}, we add additional nodes and relations between nodes to the knowledge graph. KSG has two types of nodes, entities and attributes, as well as two types of directed edges, which indicate entity-entity and entity-attribute relationships.
Specifically, entity nodes include fact nodes (such as person, etc.), environment nodes, and skill nodes. Entity description, skill display, pre-train network, and offline dataset are examples of attribute nodes.
KSG retains initial information retrieval and question-and-answer functionalities. Meanwhile, KSG can search for and display the skills of different agents in various environments, as well as give transferable knowledge for learning new skills.
If we want to train a new skill, we can search for learned skills in KSG related to the new skill as a pre-training networks and use them to train the new skill. Our experiments demonstrate that KSG can achieve effective skill searching and learning. The main contributions of this paper can be summarized as follows:

\begin{figure*}[t] 
  \centering
  \includegraphics[width=13cm]{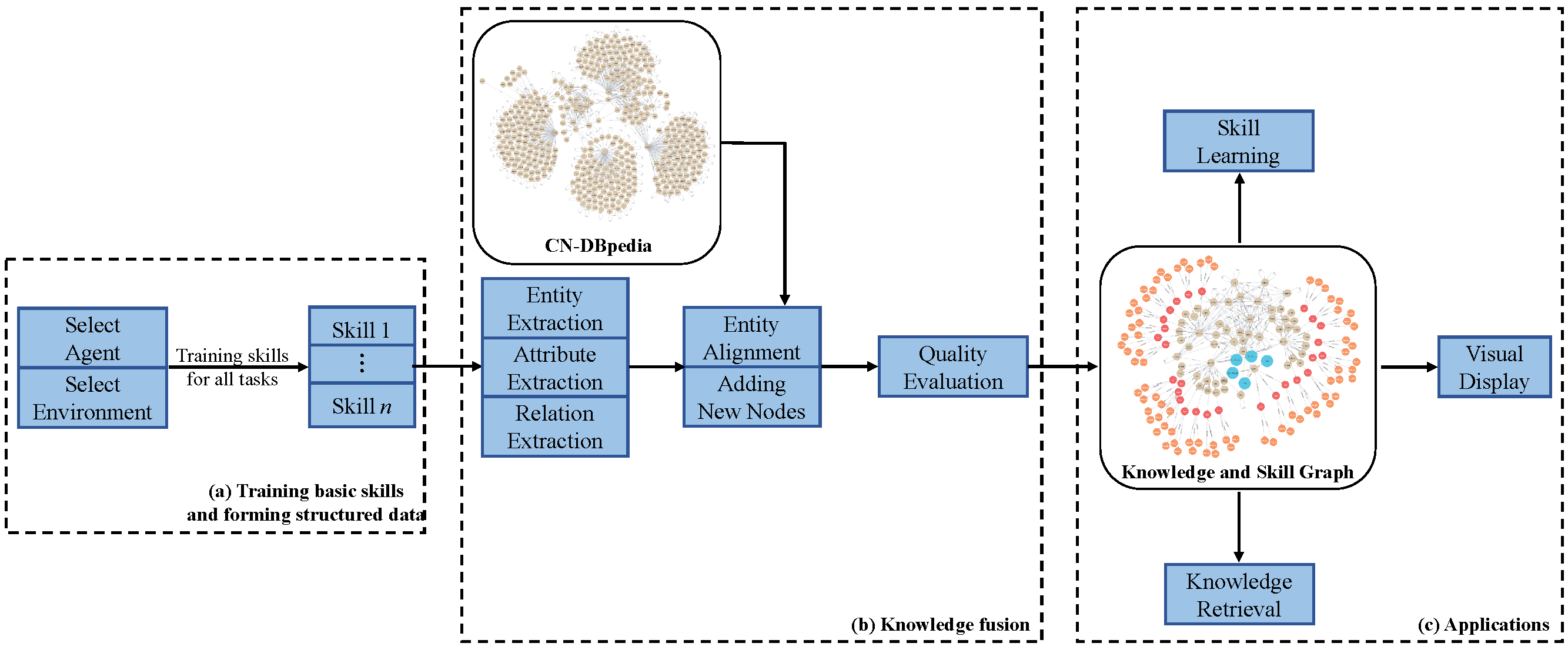}
  
  \caption{The architecture of knowledge and skill graph. We first train basic skills and form structured data in (a). Then, we extract dynamic behavioral information from basic skills and construct a preliminary but specific KSG based on CN-DBpedia in (b). Finally, we apply KSG to achieve knowledge retrieval, visual display and skill learning in (c).}
\end{figure*}

\begin{itemize}
    \item In order to extend KG to deal with dynamic behaviors, we propose a novel concept of KSG to simultaneously process static and dynamic knowledge.
    \item We propose and visualize a preliminary and specific KSG. To the best of our knowledge, our work is the first to implement skill retrieval and skill reuse by introducing behavioral intelligence into KG.
    \item Extensive experiments demonstrate the effectiveness of KSG in a variety of tasks, such as QA, knowledge retrieval and skill learning.
\end{itemize}

\section{Knowledge and Skill Graph (KSG)}

Our aim is to establish a knowledge and skill graph (KSG) that not only retains the original static facts of the knowledge graph but also deals with dynamic behavior information. As a result, KSG can be utilized for both text-based Q\&A systems, retrieval and recommendation, and skill retrieval, as well as providing transferable knowledge for learning new skills. In this paper, we emphasize the last functionality.
As shown in Figure 1, knowledge and skill graph is constructed based on CN-DBpedia and the construct process is divided into three parts: a) Training Skills for Data Preparation; b) knowledge Fusion; c) Applications. We first introduce the methods and process of training basic skills and the storage of skills. Then, we elaborate on the details of knowledge fusion. Finally, we explain the main applications and the details are described in section 3.

\subsection{Training Skills for Data Preparation}
In the processing of constructing KSG, we firstly train a great deal of basic skills of different agents in different environments. In this paper, the agent Humanoid, Ant and Half Cheetah are from Mujoco \cite{mujoco2012}. Environment plane is smooth ground and obstacle represents an obstructed ground. Then, we design different tasks according to the direction of walk. Finally, we consider the 18-DoF quadruped robot. For simulative quadruped robot, we design three environments: plane, stair, and irregular plane, and only consider the walk task. Specifically, we also store the walk skill of real quadruped robot in the environment plane.
In this paper, we use SAC \cite{SAC2018} to train basic skills. By choosing different tasks, agents and environments, we have trained 23 different skills.
For each skill, we store such trained network and offline data as knowledge which will be used to construct KSG.

\subsection{Knowledge Fusion}
In this paper, our preliminary KSG is based on the CN-DBpedia dataset. During the process of the KSG construction, we use neo4j and py2neo to import the triple relationship of CN-DBpedia. 
Among them, Neo4j is a powerful graph database tool that can search specific nodes with O(1) time complexity, and Py2neo is a python package that allows us to manipulate the neo4j database in a python environment. 
Once we have imported the CN-Dbpedia database, we can directly build the KSG based on it by adding a variety of skills, environments and agents.
Different from traditional KG, we innovatively introduce dynamic behavior information into KG and add new nodes where added entity nodes consist of agents, environments and skills, as well as attribute nodes include entity description, skill display, pre-trained network and offline dataset.

To obtain new nodes about skills, we first need knowledge extraction from trained basic skills. In this paper, the knowledge extraction is divided into three types: entity extraction, attribute extraction and relation extraction. For entity extraction, we use named-entity recognition (NER) to classify entity into pre-defined categories such as human, plane, walk up, walk right and so on. In this part, we extract new entity nodes including agent, environment and skill.
After entity extraction, the attribute extraction is to define the attribute or description of entity. For relation extraction, we use rule-based and dictionary-based methods to extract the relationships among the entities and attributes. Relation extraction is to find the relations between entity-entity and entity-attribute. 
For example, \textit{``huanmoid''} is one of the most popular agents which have been broadly used in various Reinforcement learning researches, and has many human's features, so we use NER to find entity node \textit{``human''} in CN-Dbpedia, and then add some pre-defined skill entity nodes (such as \textit{``Walk\_Up''}), environment entity nodes (\textit{``Plane''}, \textit{``Obstacle''}) to it.

\subsection{Applications}
Except for the construction of KSG, we also build the KSG Question-Answer system (KSGQA) based on the KSG to facilitate the use of prior knowledge (in this paper, prior knowledge means the trained models as well as identity representations). This system can interpret and encode the specified queries in order to acquire target triplets.
As shown in Figure 2, our KSGQA utilizes fine-tuned Bert \cite{2018bert} to encode query. In this way, we can identify entity and relations between entities from the query. According to the entity, we can first obtain the corresponding triplets including entity, relation and attribute from KSG, and then we use the relation to determine the target triplets. 
For example, when inputting a query \textit{``Do you know what skill humans have?''}, we can use Bert to extract entity \textit{``human''} and \textit{``skill''}. Meanwhile, the relation \textit{``have''} is used to obtain target triplets such as \textit{``\{`human', `have', `walk\_left'\}''}, \textit{``\{`human', `have', `walk\_right'\}''} and so on. Based on the KSGQA, we can realize some functionalities such as knowledge retrieval, visual display and skill learning.

\begin{figure}[t] 
  \centering
  \includegraphics[width=5cm]{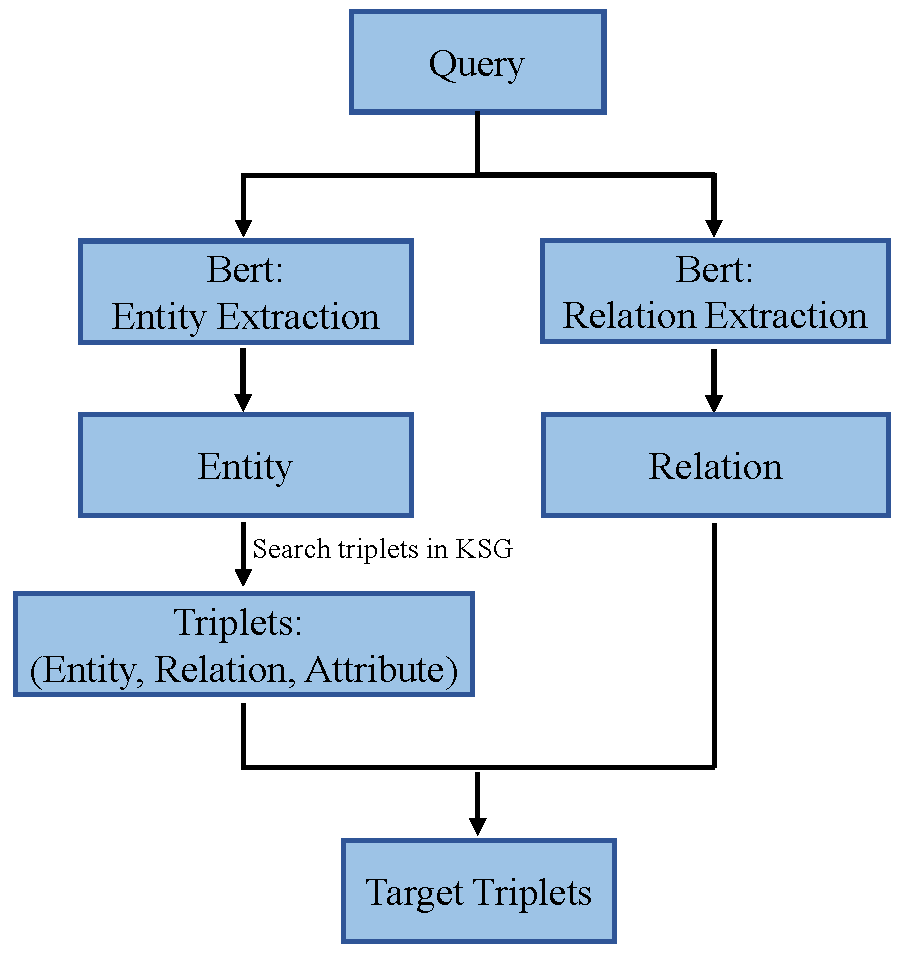}
  \caption{Process of KSGQA}
\end{figure}

For knowledge retrieval, we can retrieve those skills that an agent has learned and those data to be called. After we obtain a list of the agent's skills, we can display those skills as well. For example, when we input \textit{``Do you have know what skills \{agent\_name\} have?''} and obtain a list of the agent's skills, we can input query \textit{``Can you show \{agent\_name\} \{skill\_name\} in the \{environment\_name\}?''} to load the video and show such skill. In addition, we store corresponding trained model into KSG so that we just have to query this KSGQA system to load the trained model. For example, when we input \textit{``Can you search \{agent\_name\} \{skill\_name\} in the \{environment\_name\}?''}, we can get the trained model and offline data of the agent. Thus, if we need a new skill that is not available in KSG, the trained model and offline data can be regarded as pre-trained model and training data to learn new skill \cite{chen2021deep,chen2021pareto,xiao2021learning,chen2021multi}. In this paper, we define two selection strategies for pre-training models based on environmental differences and task differences. When performing the same task in different environments, we choose the skill model with the highest similarity as the pre-training model according to the environment similarity which is determined by calculating the Euclidean distance between the sampled states in different environments. When performing different tasks in the same environment, we select the pre-training model according to the task similarity. 

\section{Experiments and Applications of KSG}

We aim to establish a KSG which can be used to search skills and provide transferable knowledge for learning new skills. In this section, we first show KSG's Q\&A system and skill retrieval capabilities. Then, we use KSG to retrieve and provide related skill model and offline data for learning new skills. 

\subsection{Knowledge retrieval and display}

In this paper, we construct a preliminary and specific KSG based on CN-DBpedia. Compared with traditional knowledge graph, KSG focuses on behavioral intelligence. We retain the original functions of the knowledge graph, meanwhile introducing new dynamic knowledge and skills into it. Therefore, we also design a corresponding knowledge and skill graph question answer system. The important functions of KSGQA are knowledge retrieval and display. 
As shown in Figure 3, we show the basic functions of knowledge retrieval.
We can see that KSG can be used to retrieval existing skills as shown in Figure 3 (a), and then when we need the networks and offline data of these skills, we can call and download them from KSG in Figure 3 (b). In addition, KSG can display the stored skills. For example, when we input \textit{``Can you show ant walking down in the plane?''}, the KSGQA will load the video and show how the ant walks toward the down.
Moreover, if we need a new skill which is not existing in KSG, KSG is able to quickly select most related base skills as pre-trained models to help learn new skills. 

\begin{figure}[h] 
  \centering
  \subfigure[Skill retrieval]{
      \includegraphics[width=5.5cm]{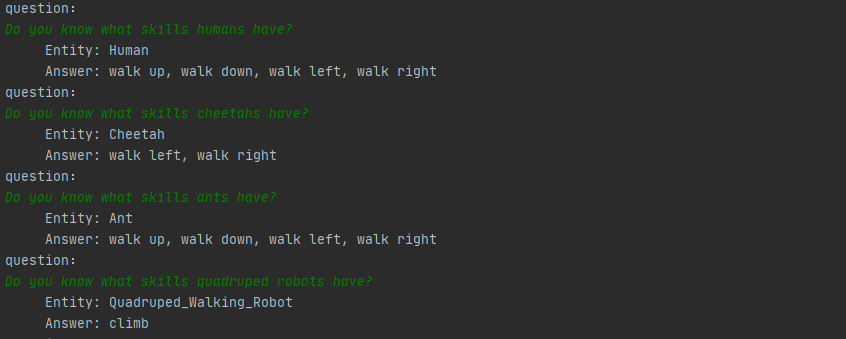}
  }
  \subfigure[Pre-trained model retrieval]{
      \includegraphics[width=5.5cm]{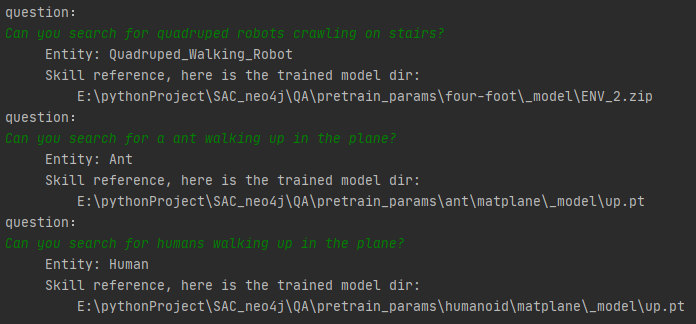}
  }
  \caption{KSGQA for knowledge retrieval.}
\end{figure}

  

\subsection{Skill Storage and New Skill Learning}
In this paper, we consider agents from Mujoco and real quadruped robot as shown in Figure 4. Each agent has different skills in different environments. We add skill nodes and corresponding attribute nodes to store these skills in KSG, where the stored knowledge includes pre-trained model, video and offline data. 

\begin{figure}[thb] 
  \centering
  \subfigure[Humanoid]{
      \includegraphics[width=3.5cm]{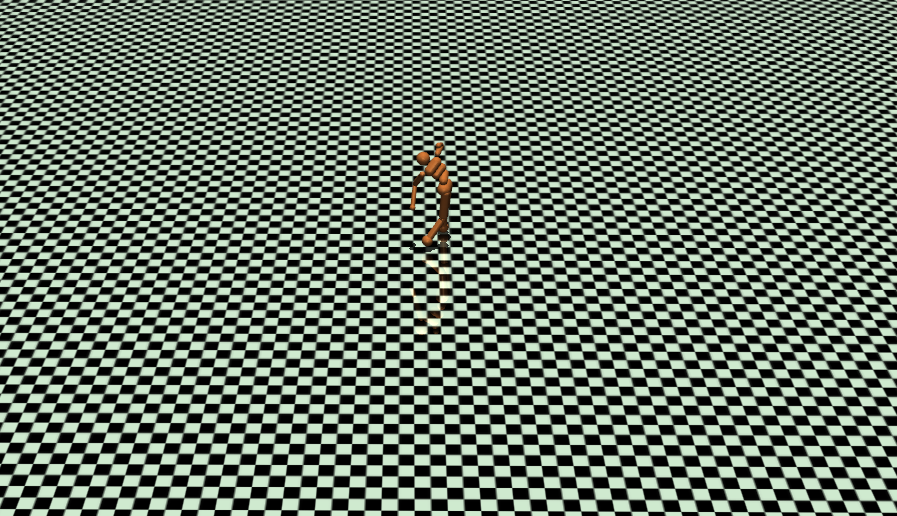}
  }
  \subfigure[Ant]{
      \includegraphics[width=3.5cm]{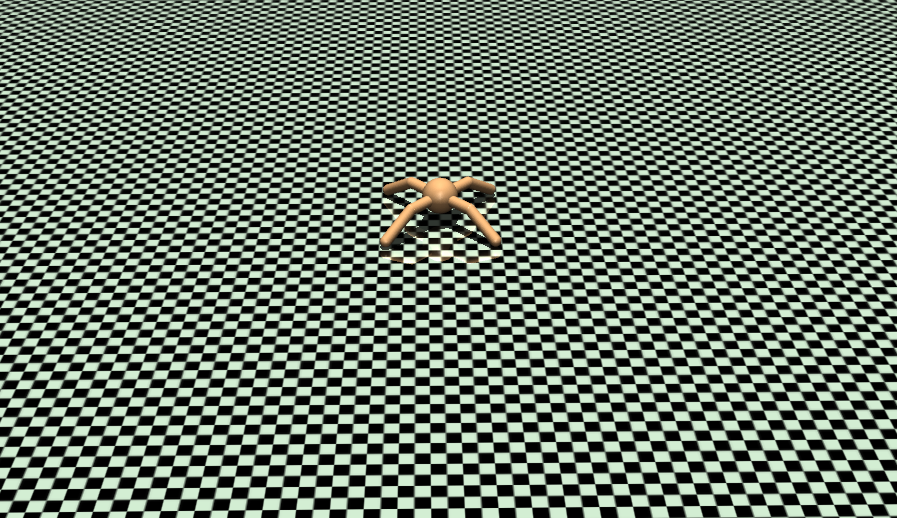}
  }
  \subfigure[Half cheetah]{
      \includegraphics[width=3.5cm]{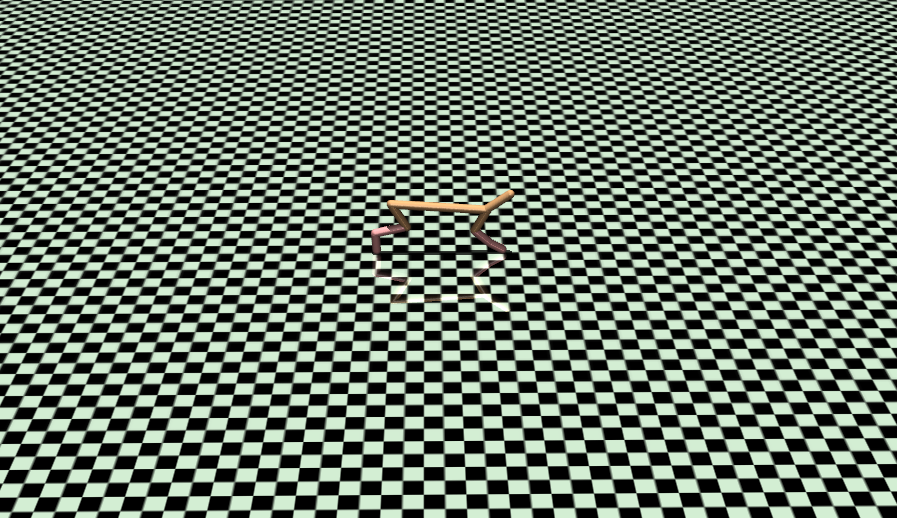}
  }
  \subfigure[Quadruped Robot]{
      \includegraphics[width=3.5cm]{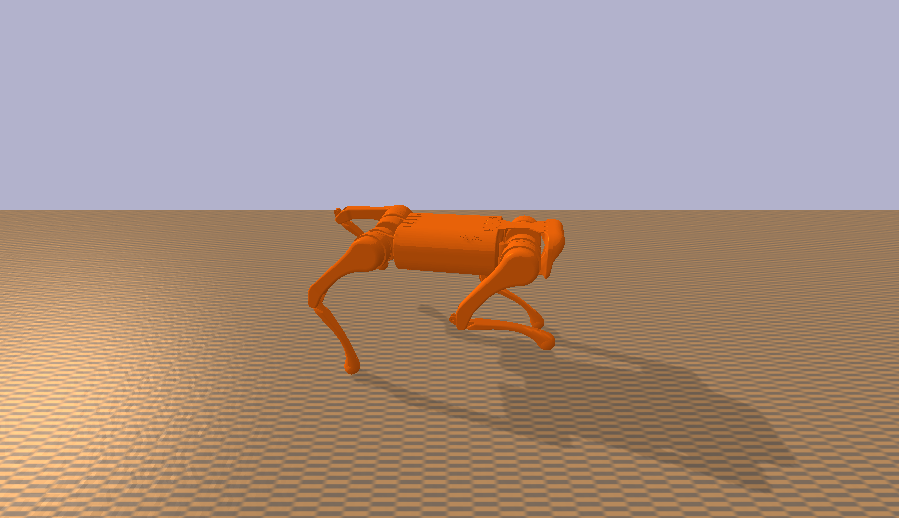}
  }
  \caption{All agents being considered in our specific KSG.}
\end{figure}

\begin{figure}[thb] 
  \centering
  \subfigure[Plane]{
      \includegraphics[width=2.5cm]{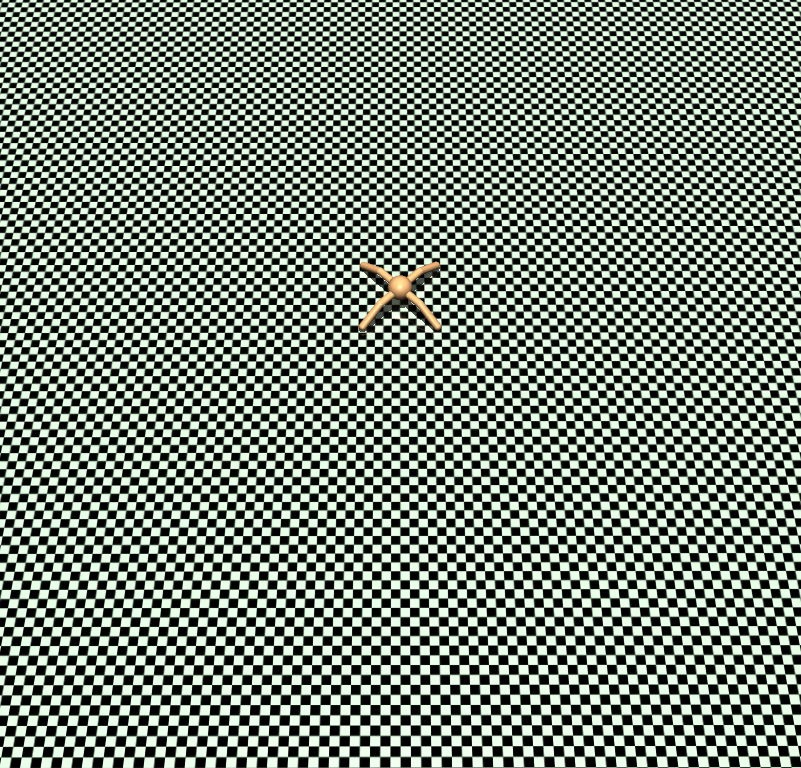}
  }
  \subfigure[obstacle]{
      \includegraphics[width=2.5cm]{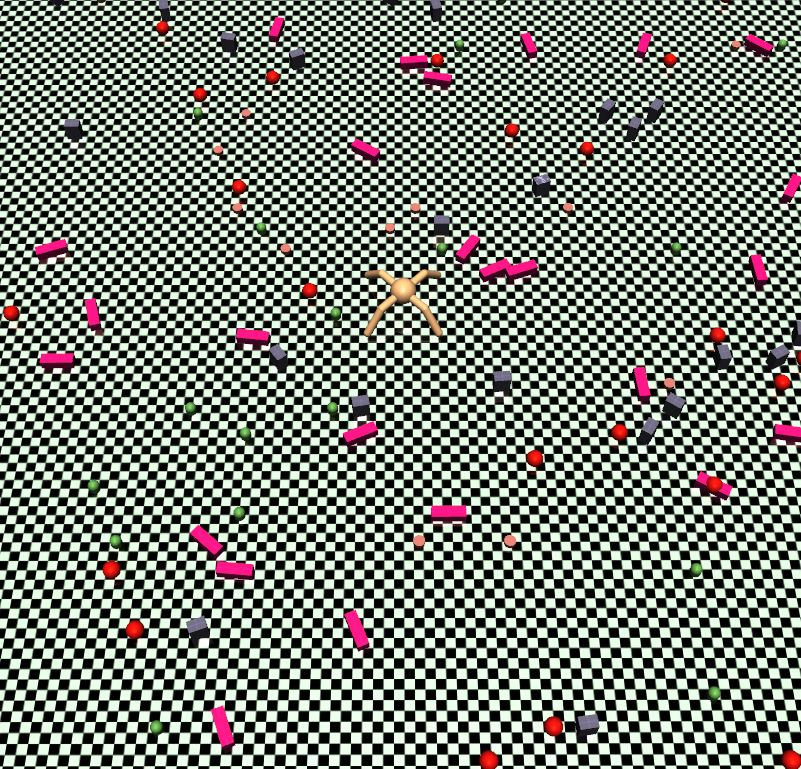}
  }
  \subfigure[stair]{
      \includegraphics[width=2.5cm]{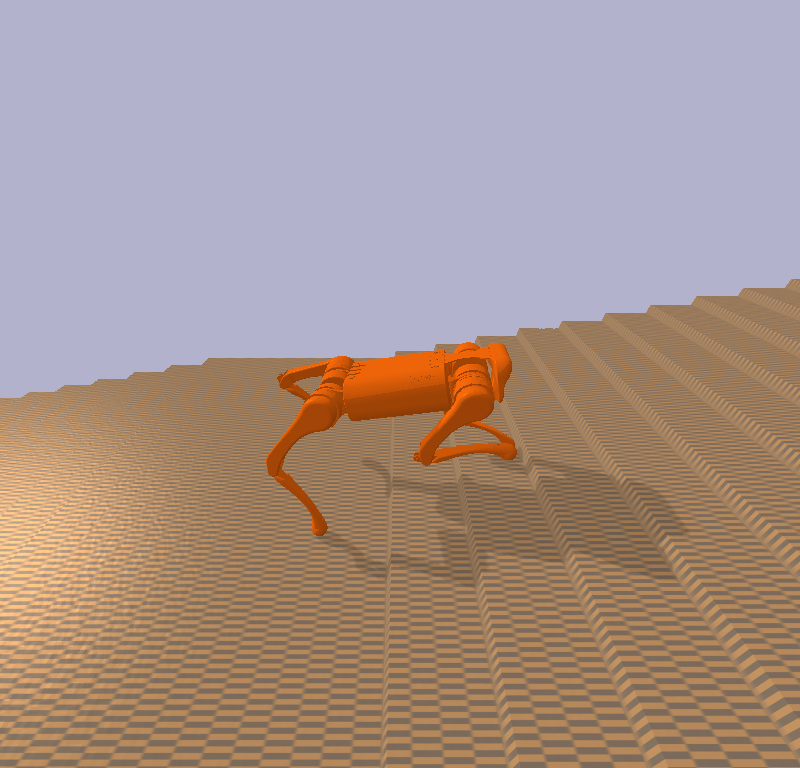}
  }
  \caption{All three environments in our preliminary but specific KSG including plane, obstacle and stair.}
\end{figure}

In order to enrich our KSG, we design different environments to complete each task. A part of these environments are shown in Figure 5, including \textit{Plane}, \textit{obstacle}, and \textit{stair}. 
By performing  different tasks in different environments, we can acquire different skills for each agent. These stored skills are transferable knowledge that can be used to learn new skills.
In this part, we use KSG to help agent learn new skills in different environment. Actually, the transferable knowledge includes pre-trained neural network and offline dataset. Therefore, we can directly use one of the most relevant models as a pre-training model for new skills, or we can use multiple related skills to combine and learn a new skill. 

\begin{figure}[hb] 
  \centering
  \subfigure[Quadruped Robot]{
      \includegraphics[width=4cm]{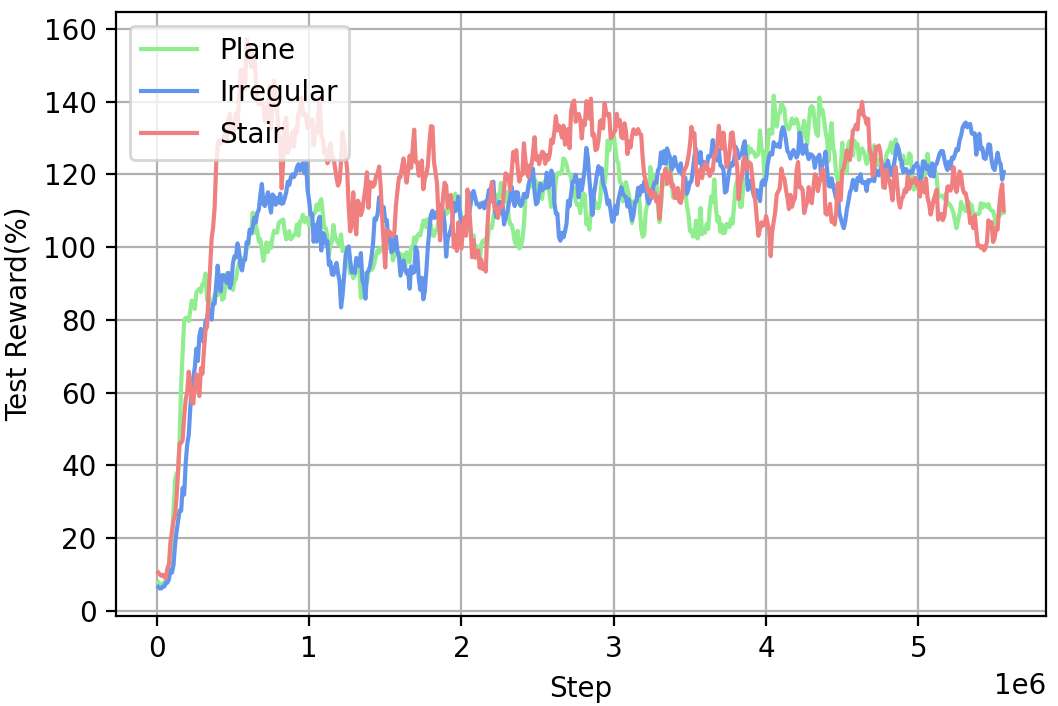}
  }
  \subfigure[Walk in irregular]{
      \includegraphics[width=4cm]{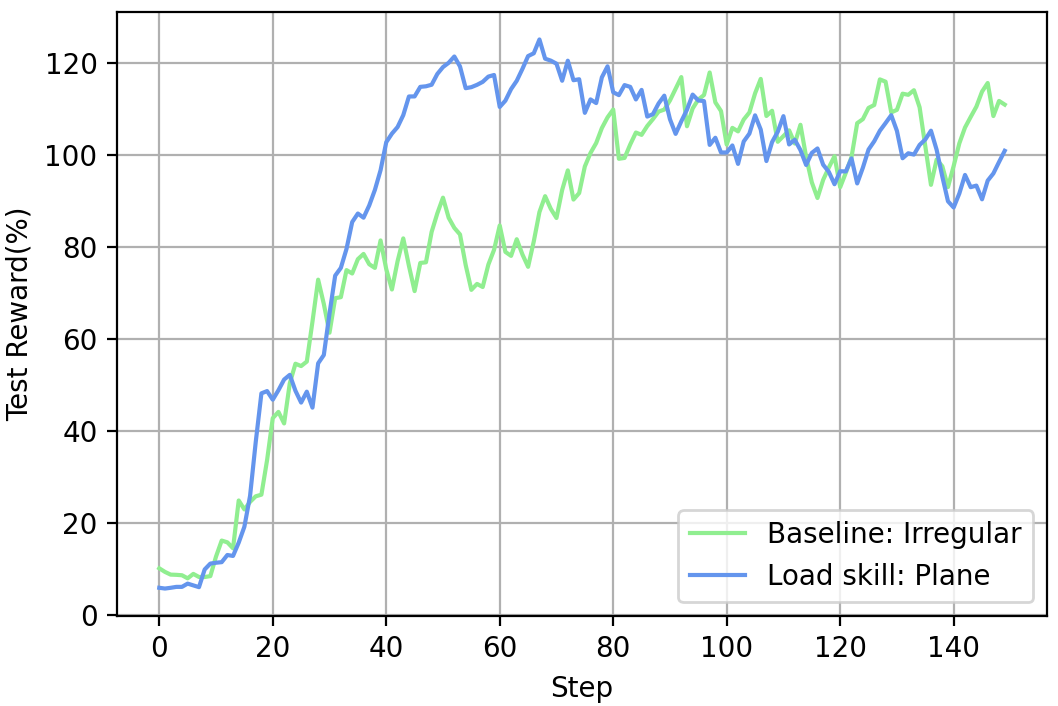}
  }

  \caption{Test reward of Quadruped Robot and learn new skills in new environment using pre-trained model.}
\end{figure}

In this paper, we can load stored skill model from KSG as pre-train model to learn new skill. In Figure 6 (a), we show the stored skill model's test reward of real Quadruped Robot. If we now need to acquire the skills of walk for quadruped robot in environment irregular, we can select the most relevant skills from KSG as a pre-training model. This problem belongs to performing the same task in different environments. We first calculate the similarity of environment between irregular, plane and stair. In the above question, we select the skills with the highest similarity by calculating task similarity (Walk in environment plane) as the pre-training model to learn new skill walk in environment irregular. As show in Figure 6 (b), we can see that loading related pre-training models can improve training efficiency and reduce nearly half of the training time compared with direct training from scratch.


\begin{figure}[h] 
  \centering
  \subfigure[30 degrees]{
      \includegraphics[width=4cm]{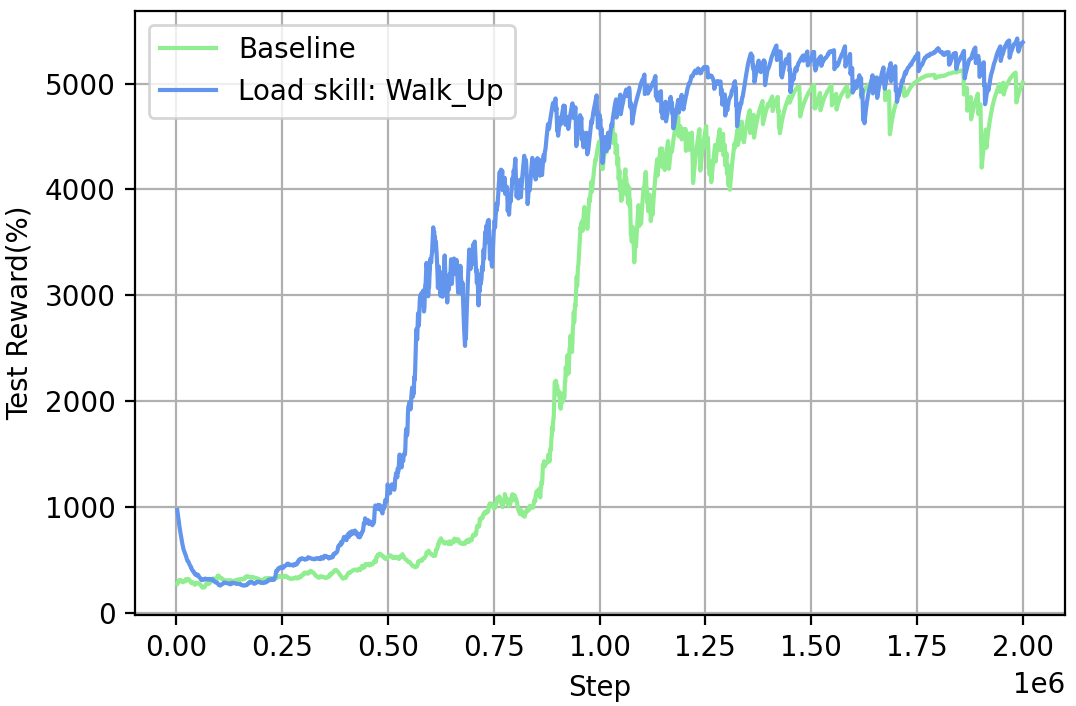}
  }
  \subfigure[60 degrees]{
      \includegraphics[width=4cm]{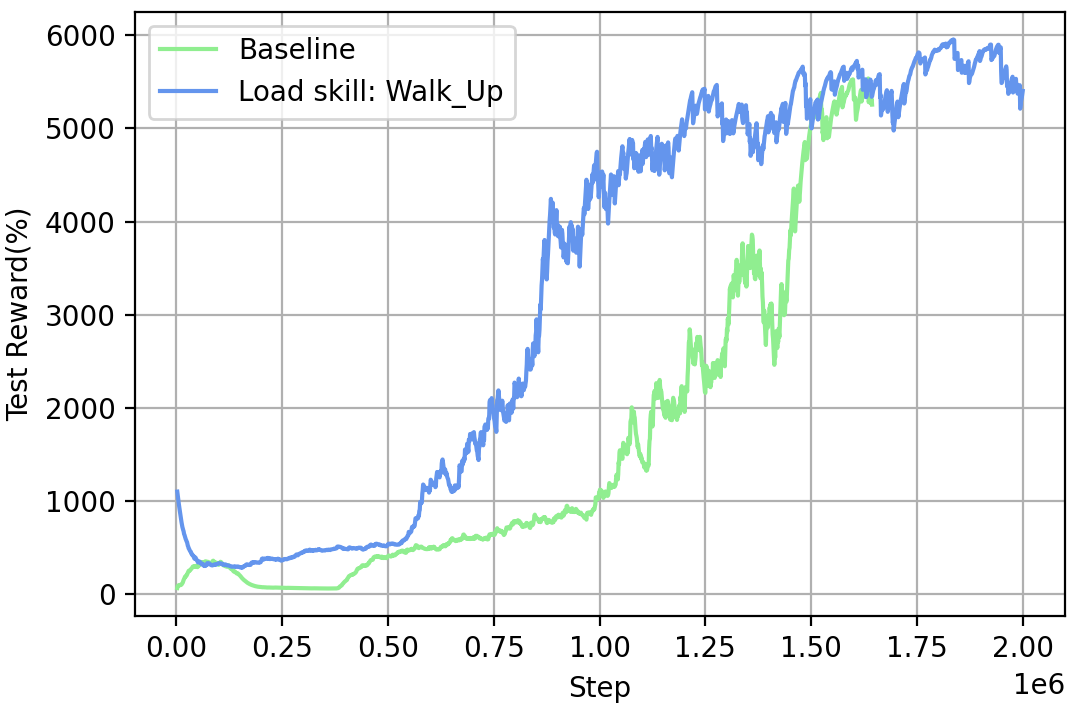}
  }
  \caption{Obliquely upward walk at a different Angle from the horizontal direction}
\end{figure}

On the other hand, if we need a new skill which is performing different tasks in the same environment, we can select the pre-training model according to the task similarity. When we need two skills that are obliquely upward walk at a different angle from the horizontal direction (30 degrees, 60 degrees), we can automatically select skill walk right and walk up as pre-training models according to the task similarity. As shown in figure 7, we can observe that loading pre-training skills according to task similarity can also improve the learning efficiency of new skills.

\section{Conclusion and Future Work}

In this paper, we propose a novel concept of KSG to simultaneously process static and dynamic knowledge. KSG introduces dynamic behavioral information to KG, which can also implement skill retrieval and skill reuse.\
Existing methods on KG usually focus on text pro-cessing and static fact, so we extend KG to deal with dynamic behaviors.
KSG retains the original function of the knowledge graph and adds new functions including skill retrieval, visual display of skill and knowledge transferring for new skill learning. In experiments, we show all of KSG's capabilities in detail. It indicates that constructing KSG is significant and valuable for storing and utilizing dynamic behavioral information. 

Although our KSG now is preliminary, it is very meaningful. In the future, we will continue to expand and improve KSG for real applications. On the one hand, KSG will provide a large amount of basic skills and offline data for reinforcement learning, meta learning, imitation learning and so on. 
On the other hand, KSG will also provide more complex relation between skills, agents and environments for skill learning and reasoning.

\newpage
\bibliographystyle{ACM-Reference-Format}
\balance
\bibliography{reference}

\end{document}